\documentclass[runningheads]{llncs}
\usepackage{graphicx}
\usepackage{hyperref}
\hypersetup{colorlinks=true,linkcolor=blue,urlcolor=blue}
\newcommand{\etal}{\textit{et al.}}

\begin{document}
%
\title{Versailles-FP dataset: Wall Detection in Ancient Floor Plans 
}
%
\author{Wassim Swaileh\inst{1} \and
Dimitrios Kotzinos\inst{1} \and
Suman Ghosh\inst{3}\and
Michel Jordan\inst{1}\and \\
Son Vu\inst{1}\and
Yaguan Qian\inst{2}}
\authorrunning{W. Swaileh et al.}
%
\institute{ETIS Laboratory UMR 8051, CY Cergy Paris University, ENSEA, CNRS, France \email{firstname.lastname@ensea.fr}
\and
Zhejiang University of sciences and technologies, China\\
\email{qianyaguan@zust.edu.cn}
\and
RACE, UKAEA, United Kingdom \\
\email{suman.ghosh@ukaea.uk}}
%
\maketitle              
\begin{abstract}
Access to 
historical monuments’ floor plans over a time period 
is necessary to understand the architectural evolution and history. Such knowledge bases also helps to rebuild the history by establishing connection between different event, person and facts which are once part of the buildings. 
Since the two-dimensional plans do not capture the entire space, 3D modeling sheds new light on the reading of these unique archives and thus opens up great perspectives for understanding the ancient states of the monument.
Since the first step in the building's / monument’s 3D model is the wall detection in the floor plan, we introduce in this paper the new and unique Versailles-FP dataset of wall groundtruthed images of the Versailles Palace dated between $17^{th}$ and $18^{th}$ century.
The dataset's wall masks are generated using an automatic approach based on multi-directional steerable filters. The generated wall masks are then validated and corrected manually.
We validate our approach of wall-mask generation in state-of-the-art modern datasets.
Finally we propose a U-net based convolutional framework for wall detection.
Our method achieves state-of-the-art result surpassing fully connected network based approach.


\end{abstract}
\section{Introduction}
Time and tide wait for none, however we can try to rebuild the time from the evidence that is present and depicts the stories of the past. Historical documents about ancient architecture provide important evidence regarding the time, planning, design and understanding of scientific methods practiced.
These documents can be used for different purposes like understanding and recreating history, which can be used for facilitating academic research, restoration, preservation and tourism.
Architectural plans are scaled drawings of apartments or buildings. They contain structural and semantic information; for example, room types and sizes, and the location of doors, windows, and fixtures. Modern real-estate industry uses floor plans to generate 3D viewing for prospective clients. Analogously, ancient architecture plans can also be used to create 3D visualisations for interested audiences (academics, tourists etc.). These documents not only depict the stories of a certain time but also depict the altercations and modifications throughout history, thus, enabling the audience with a timeline, where one can navigate through and get a glimpse of the past. 
Recently this type of research activities and projects have been undertaken for different cities throughout Europe \footnote{https://www.timemachine.eu/}, however they rely mainly on historical information (text documents) rather than architecture. 

One major and crucial component of floor plan analysis is wall detection as walls are the building blocks of rooms and define the structure of a building. 
Thus wall (or structure of walls) depicts the evolution of the architecture of the building and the way the different rooms started emerging and playing a specific role in the life of the structure.
Thus wall detection is one of the most crucial step in analysis of floor plans and often features as sub-step (implicit or explicit) of many other research areas involving floor plan analysis.

However, wall detection is a difficult task to automate due to the fact that walls can be easily confused with different lines existing on the ancient floor plans and thus leading to many false identifications.

\begin{figure*}[!h]
\centering
\includegraphics[width=1.0\textwidth]{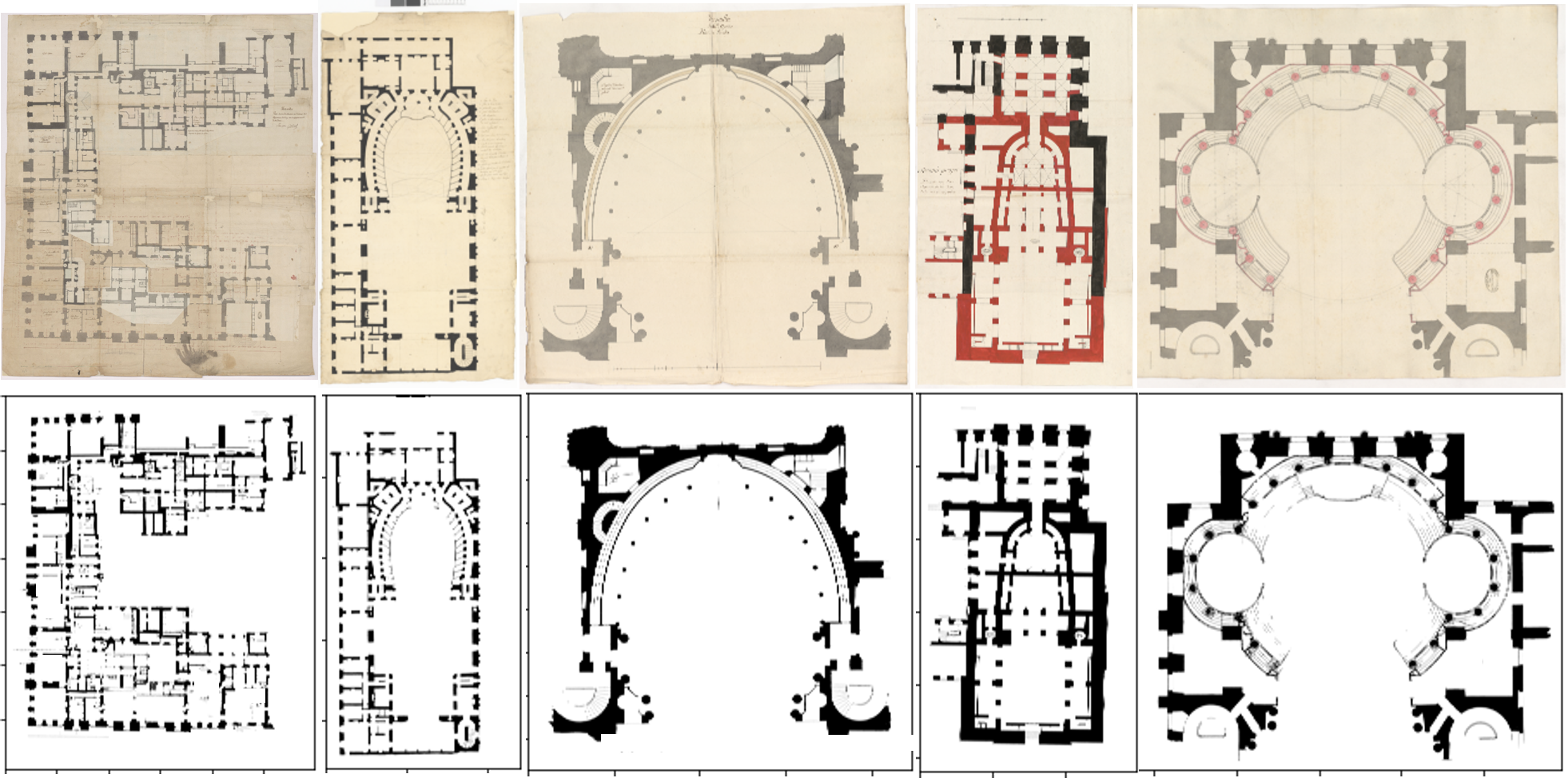}
\caption{Samples of Versailles palace floor plans dataset (Versailles-FP) annotated automatically and corrected manually. Original image and their corresponding masks are in the top and down rows of images respectively}
\label{dataset}
\end{figure*}

One of the main bottlenecks in this regard is the unavailability of floor plan documents regarding ancient architectures. In this paper, we propose a new dataset Versailles-FP to mitigate that. In particular we propose a dataset consisting of floor plans of the Versailles Palace drawn up in the period between $17^{th}$ and $18^{th}$ century. The dataset consists of 500 floor plan (FP) images associated with their corresponding wall mask ground truth. Though the dataset can be utilized in many different research problems and further extension is both planned and encouraged but for the time being we are focuses on wall detection problem. 
The dataset is available under agreement for the academic research community via the web page: \url{https://www.etis-lab.fr/versailles-fp/}.
Figure \ref{dataset}, shows samples of the Versailles-FP dataset.

There has been attempts to use document analysis and computer vision techniques to study floor plans but barring only a few \cite{tabia2016automatic} most dealt with modern architectures.

In summary our contribution can be stated as:
\begin{itemize}
    \item A novel floor plan dataset for ancient architecture (Versailles-FP)
    \item A method to automatic generation of wall masks and creation of groundtruth
    \item A CNN based wall detection pipeline and baseline results in proposed Versailles-FP dataset 
    \item Analysis and comparison of proposed wall detection and mask generation approaches with state-of-the-art techniques
\end{itemize}
The rest of the paper is organized as follows: in Section \ref{sec:related_works}, we provide a brief review of most related articles. In section \ref{sec:Dataset_Description} we provide the details of proposed dataset along with semi-automatic groundtruth generation procedure. In section \ref{sec:unet_wall}, Proposed CNN framework for wall detection is described.
Finally in section \ref{sec:dl_exper} and \ref{sec:conclusion} we provide experimental validation and conclusion respectively.


\section{Related works}
\label{sec:related_works}
Research on architectural layout documents mainly focused on modern architectures due to lack of ancient architectural documents.
In the context of modern architectural images different research problems has been proposed which involves wall detection
In the following two subsections, we briefly describe prior works involving wall detection, namely 3D reconstruction and wall detection.


\subsection{Automatic 3D model generation from floor plans}

Tombre’s group was one of the pioneer in this field. In \cite{Tombre1}, they tackle the problem of floor plan 3D reconstruction. They continued their research in this direction in \cite{dosch1999reconstruction} and \cite{dosch2000complete}. In these works, they first use a prepossessing step on paper printed drawings to separate text and graphics information. Subsequently a graphical layer separates thick and thin lines and vectorizes the floor plan. Thick lines are considered source of walls whereas the rest of the symbols (e.g. windows and doors) originate from thin lines.
Different geometric features are used to detect walls, doors and windows 
For example doors are sought by detecting arcs, windows by finding small loops, and rooms are composed by even bigger loops.
At the end, 3D reconstruction of a single level \cite{Tombre1} is performed. This is extended in \cite{dosch1999reconstruction} and \cite{dosch2000complete} by learning correspondence of several floors of the same building by finding special symbols as staircases, pipes, and bearing walls. They also observe the need of human feedback when dealing with complex plans in \cite{dosch1999reconstruction} and \cite{dosch2000complete}. Moreover, the symbol detection strategies implemented, are oriented to one specific notation, thus clearly insufficient for ancient architecture, where plans are consistently changing for 100's of years.

Similarly, Or \etal in \cite{or2005highly} focus on 3D model generation from a 2D plan. However, they use QGAR tools \cite{rendek2004search} to separate graphics from text. One major disadvantage of their work is the need of manual interactions to delete the graphical symbols as cupboards, sinks, etc. and other lines which can disturb the detection of the plan structure. 
Once only lines belonging to walls, doors, and windows remain, a set of polygons is generated using each polyline of the vectorized image. At the end, each polygon represents a particular block. Similar to  \cite{dosch1999reconstruction} Or et al. also used geometric features to detect doors, windows etc.
This system is able to generate a 3D model of one-stage buildings for plans of a predefined notation.
Again this poses a serious problem in case of ancient architectural plans where the notation styles are varied in nature.

More recently Taiba \etal \cite{tabia2016automatic} proposed a method for 3D reconstruction from 2D drawings for ancient architectures. There approach also follows similar pipelines, more specifically they used specially designed morphological operations for wall detection tasks and they joined different walls to form a graph which then utilized to form the 3D visualizations. However their method is not fully automatic and need to be parameterize for every types of floor plan.   

\subsection{Wall Detection}


Similar to other tasks, wall detection also relies traditionally on a sequence of low level image processing techniques including preprocessing to de-noise the floor plan image, text/graphics separation, symbol recognition and vectorization.

Ryall et al. \cite{ryall1995semi} applied a semi-automatic method for room segmentation.
 Ma{´}ce \etal uses Hough transform \cite{mace2010system} to detect walls and doors. Subsequently Wall polygons are created by the hough lines and are partitioned iteratively into rooms, assuming convex room shapes. Ahmed et al. \cite{ahmed2012automatic} process high-resolution images by segmenting lines according to their thickness, followed by geometrical reasoning to segment rooms. Doors are detected using SURF descriptors. 
 More recently researchers adopted convolutional neural networks e.g. Dodge \etal in \cite{DodgeMVA2017} used a simple neural network for the removal of noise followed by a fully convolutional neural network (FCN) for wall segmentation.
 Liu et al. \cite{liu2017raster} used a combination of deep neural networks and integer programming, where they ﬁrst identify junction points in a given ﬂoor plan image, and then join the junctions to locate the walls in the ﬂoor plan. The method can only handle walls that align with the two principal axes in the ﬂoor plan image. Hence, it can recognize layouts with only rectangular rooms and walls of uniform thickness, this is a critical limitation in the context of historical layouts where rooms shapes are not always rectangular and often round shaped rooms are used.
  In \cite{yang2019dula}, also trained a FCN to label the pixels in a ﬂoor plan with several classes. The classiﬁed pixels formed a graph model and were used to retrieve houses of similar structures.

  
  As stated earlier all these works use different methods ranging from heuristics to geometric features, from low level image processing to deep learning and obtained noteworthy success.
  But, the efficacy/success of these techniques is a subject of investigation for ancient architectural documents. This concern is further emphasized due to the presence of noise, low differences between the FP image background and foreground colors, image size can vary between very big and very small image size, etc. In Section \ref{sec:wall_detection_challenges}, we provide a detailed discussion on the challenges posed by this.

\section{The Versailles-FP dataset}
\label{sec:Dataset_Description}
In 2013 the research project VERSPERA\footnote{\url{https://verspera.hypotheses.org/}} was started  with the aim of digitizing a large amount of graphical documents related to the construction of the Versailles palace during the $17^{th}$ and $18^{th}$ centuries. There is a large corpus of floor plans including elevations and sketches present in the archives of French National Archives. The total number of documents in this archive is around 6500 among which about 1500 are floor plans. An ambitious project to digitize this varied corpus is taken in 2017, extraordinary technical capability is needed to achieve this task due to fragile and varied nature of the paper documents (for example some document can be as big as 3m $\times$ 4m). 
This project helped to provide free digital access to these stored knowledge, which were bind in paper form. This digital access open enormous possibilities for further research in various domains from history. digital humanity, architecture, computer vision, document analysis etc.
The idea of VERSPERA project can also be extended to other monuments of historical importance in the same way.

The digitized plans of Versailles Palace  consists of graphics that illustrate the building architecture such as walls, stairs, halls, king rooms, royal apartments, etc. in addition to texts and decorations. Since Versailles Palace digitized floor plans cover 120 years (1670-1790) of architectural design, different drawing styles clearly appear in the corpus. The obtained document images open new challenging fields of research for the scientific community in document analysis.
Possible field of research can be driven for printed and handwritten text localization and recognition. 
Another interesting research direction can be thought of regarding automatic understanding of small parts of the building plans in a complete view of Versailles palace floors.

Automatic plan change detection which can utilize floor plan over a time period among other sources to establish different changes in building which took place and which didn't( but planned) is another interesting research direction. Similarly floor plan captioning or question answering tasks can be thought of which will involve visual reasoning and understanding of different knowledge sources.
Thus, different researchers or research groups can utilize this dataset for different academic purposes, but among them wall detection is one of the primary task which needs to be solved (either implicitly or explicitly) in order to tackle any of these tasks.
Thus, in this paper we design our dataset primarily focused on wall detection for ancient architectural floor plans. For constructing the Versailles-FP dataset, we selected 500 images from the available digital archive of Versailles palace floor plans.

In the following sections we first describe different different unique challenges posed by ancient floor plans (Section \ref{sec:wall_detection_challenges})
Then briefly define the wall detection task (Section \ref{sec:wall_detection_task}) and describe how we generate groundtruth using a semi-automatic procedure for Versallies-FP dataset (Section \ref{sec:maskgen})


\subsection{Challenges in wall detection vis-a-vis ancient floor plans}
\label{sec:wall_detection_challenges}
The wall detection task poses several challenges due to similar form, shape and nature of the walls with other drawings in floor plans.
Furthermore,
background color of the enclosed spaces sometimes become quite similar to the walls (for example in R-FP dataset), thus become difficult to separate using simple image processing techniques.
Compared to modern floor plans, the ancient ones poses another set of problems. Firstly they are hand made/written and does not follow any specific uniform style or format. In addition, the drawing paper sheets varies wildly in size leading to a great deal of difficulty in case of training (specialy in convolutional neural networks)
\begin{figure*}[!h]
\centering
\includegraphics[width=1.0\textwidth]{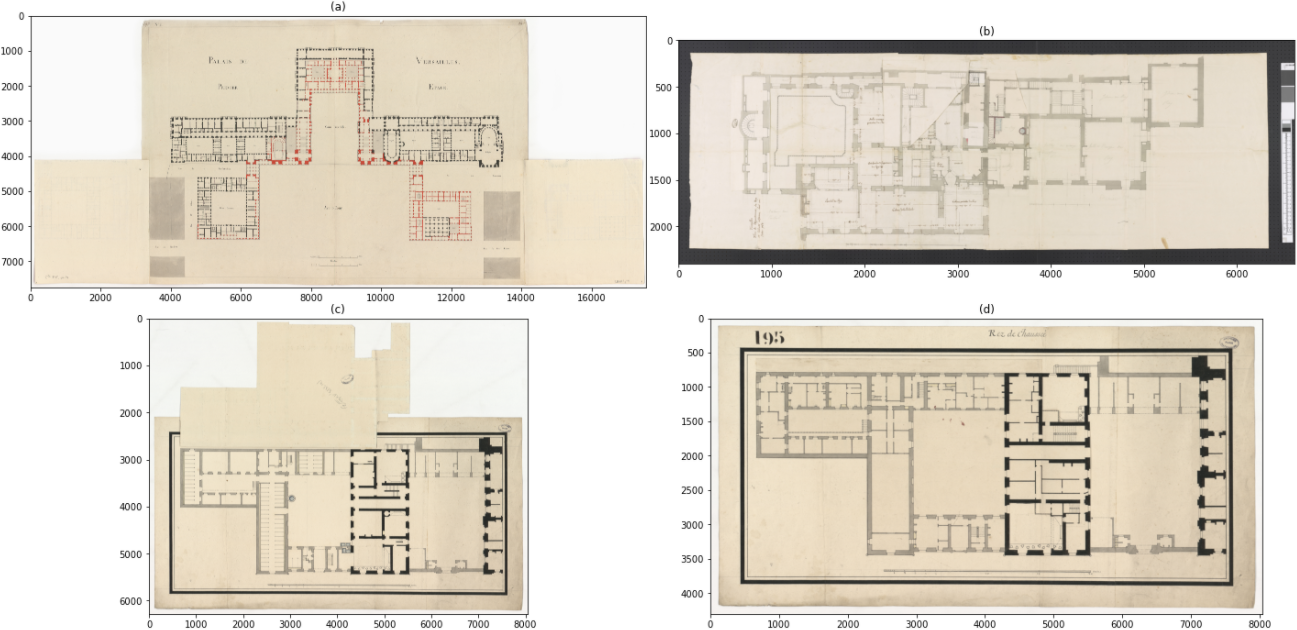}
\caption{Challenges posed by Versailles-FP (a) Black blocks represent existing walls and red ones represents planed ones; (b) Example of luminance problem (c \& d) sample of overlapped floor plans}
\label{samples}
\end{figure*}

Moreover, scanning process of such old documents add some additional noise. 
Sometimes modification to a plan is introduced by overwriting the original plan complicating the automatic processing even further.
Figure ~\ref{samples} shows some typical samples of Versailles palace floor plan images and difficulties with regard to wall detection; (a) first floor FP image where the black blocks represent existing walls and red ones represent planed ones; (b) illustrate the capturing image luminance problem with black background border; (c \& d) sample of overlapped floor plans.


\subsection{Wall detection task}
\label{sec:wall_detection_task}
In this section, we define the wall detection tasks and discuss specifics of wall to be detected throughout this paper. Later we also discuss some challenges of wall detection both in case of modern and specially for ancient floor plans.
\begin{figure*}[!h]
\centering
\includegraphics[width=0.8\textwidth]{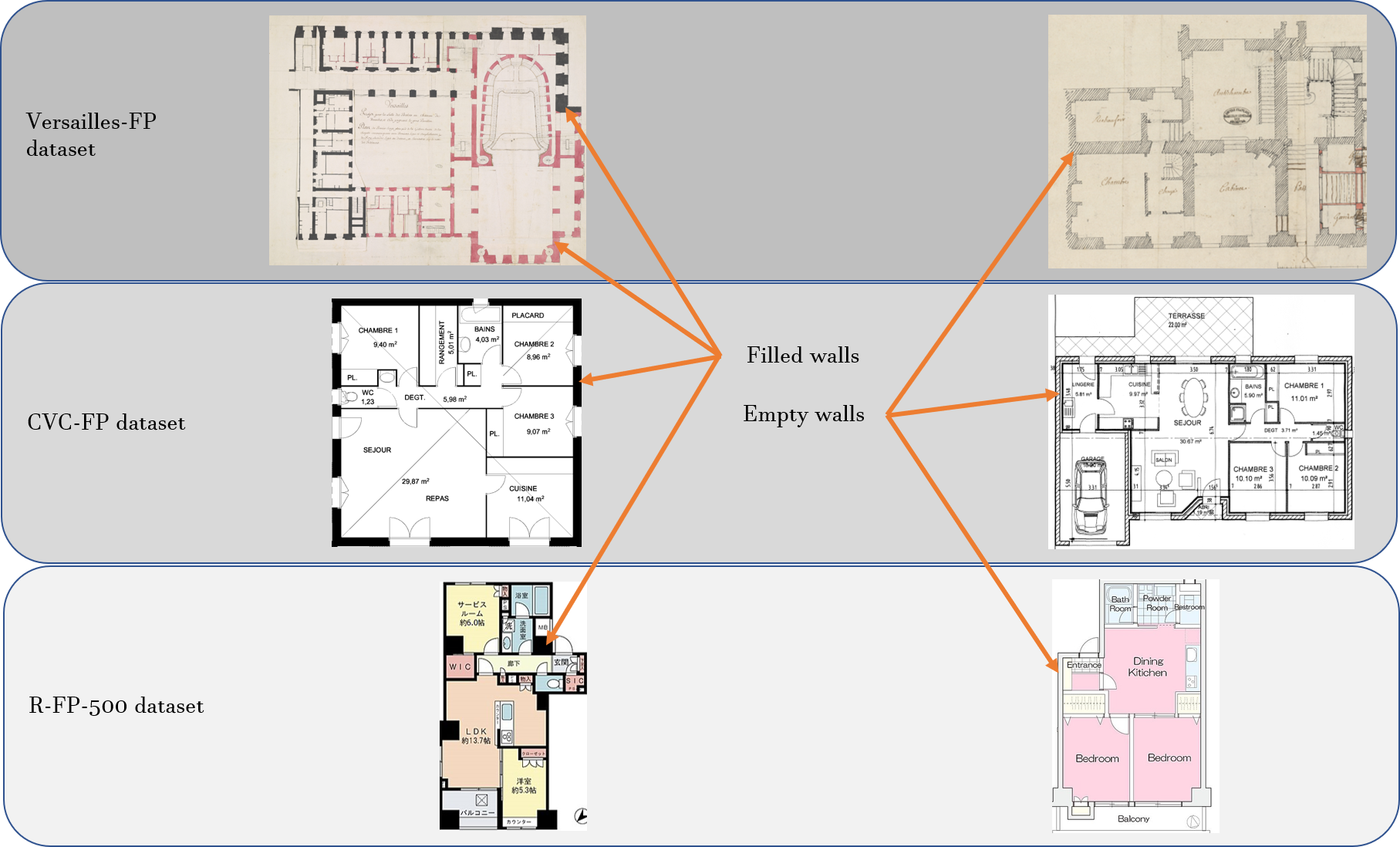}
\caption{Samples of filled and empty wall drawings (top) Versailles palace dataset (ancient), (middle) CVC-FP dataset and (down) R-FP-500 modern datasets.}
\label{wallTypes}
\end{figure*}
In the wall detection task, the aim is to identify the set of pixels corresponding to the walls. Depending on the drawing style of architects and types of walls, the representation of walls can vary. In \cite{de2015cvc}. authors has identified four different types of walls in modern floor plans, however by analysing ancient floor plans we have seen that two types of walls are sufficient to correctly identify most walls.
Thus following two types of wall drawings are taken into consideration 1) Filled wall drawing arising from thick and dark strokes with straight and rectangular form in general. 2) Empty (Hollow) wall drawing arising from twins of parallel, thin and dark strokes and sometimes connected with diagonal lines. We use this consideration for both ancient and modern drawings as our main goal is to analyse ancient floor plans. Figure~\ref{wallTypes} illustrates the two types of filled and empty wall drawings in the modern and ancient floor plans.

\subsection{Automatic wall masks generation}
\label{sec:maskgen}

For facilitating supervised learning schemes for wall detection tasks, pixel level groundtruth is needed.
Towards this goal we propose a semi-automatic procedure. First an automatic segmentation method is employed to produce binary mask corresponding to walls. Finally binary masks are checked and corrected manually to produce desired groundtruth. 
In particular, we propose a smearing segmentation method, which proved to be less sensitive to plan's drawing style variations. 
The method is inspired from ~\cite{shi2009steerable,swaileh2015multi} that target text line segmentation in text document image.

\begin{figure*}[!h]
\centering
\includegraphics[width=1.0\textwidth]{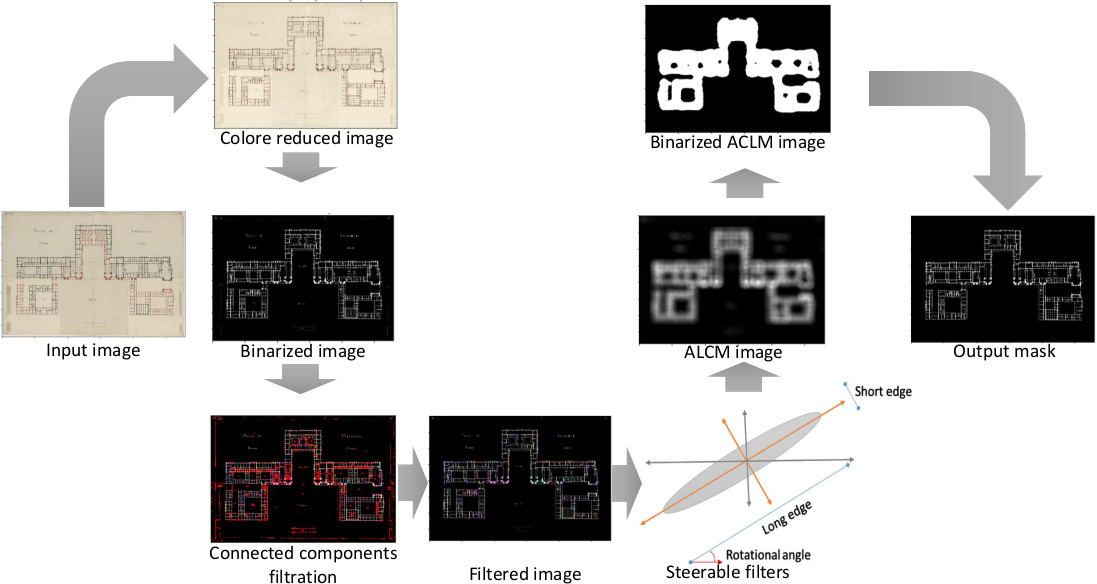}
\caption{Mask generation processing chain for Versailles palace plans' images}
\label{maskGen}
\end{figure*}

The proposed approach consists of three steps. 
First, color correction, reduction and connected component filtration processing are applied to produce noise reduced binary image of the input FP image. 
The left side column of images of figure ~\ref{maskGen} illustrates this step of the processing chain.
Next, a set of steerable filters are used in different directions to increase foreground pixel intensity and obtain the adaptive local connectivity map (ALCM) of the filtered FP image. 
Finaly, ALCM is then thresholded using Otsu global thresholding algorithm to get a fine estimation of the walls' regions. 
Theses two steps are illustrated by the right side column of images of figure~\ref{maskGen}.
The walls' mask is determined after that by superimposing the estimated wall region in the ALCM on the binary image of the plan.  

At first step, the algorithm receives colored plan image as input and corrects its contrast and luminance using exponential and gamma auto correction algorithms.
K-means classification algorithm is then used to reduce the color space in the input image from 256 RGB color into 3 RGB colors in order to enhance the quality of the grayscale of the input image. 
Next, the obtained grayscale image is binarized using Otsu thresholding algorithm.
Connected components (CC's) analysis is then conducted on the obtained binary image in order to remove small pieces of text and noise components from the image. 
Let $MDA$ and $AVA$ denote the median and the average
area respectively of the connected components of the FP image, $ECA$ denotes the estimated wall's CC area that represents a compromised value derived by $ECA = MDA+AVA$.
The idea is to turn of every CC that have area less than the estimated ECA area in the binary image as illustrated in connected components filtration image of figure ~\ref{maskGen} where 
red particles represent the removed CCs.
As a result we obtain quite clean binary image on which we conduct the operations of the second step.

In the second step, we apply a set of steerable filters \cite{shi2009steerable} in different orientation to increase the intensity of the foreground at every pixel of the filtered binary image obtained from the first step.  
A steerable filter as illustrated in figure ~\ref{maskGen} has an elliptical shape where the short edge represents the filter’s height ($FH$), the long edge represents the filter’s width ($WH$) and the rotation angle ($\theta$) represent the filter’s orientation.
According to Swaileh et al.\cite{swaileh2015multi}, the most effective parameter on the global performance of the mask generation method is $FH$. The $WH$ parameter value is indeed derived from $FH$ value, since $\theta$ affects less the performance in general.
High value of $FH$ leads to increase the intensity of all CCs including walls and unfiltered noise components, and consequently we obtain less clean walls masks. 
Empirically, We choose to set $FH$ to the previously estimated value of $ECA$.
$FW$ is derived from $FH$ by $FW = 2 \times FH$ (this setting shows best results). For effective performance we choose to use 11 filter direction values $\theta$ where $\theta \in [-25^{\circ}, 25^{\circ}]$.
The ALCM image is obtained then by applying the steerable filters with the previous configuration of $FH, WH$ and $\theta$.
At the last step, the obtained ALCM image is binarized using the Otsu thresholding algorithm. 
The final mask image is obtained by superimposing the filtered binary image and the binarized ALCM. Every pixel in the filtered binary image is turned off while its corresponding pixel in the thesholded ALCM image is off.  

This approach for generating wall masks automatically showed good performance on filled walls FP images of CVC-FP and R-FP datasets (see section \ref{sec:dl_exper}) 
However, the approach in its current version does not deal with binarization problems related to the quality or nature of the original plan image (stamps, graphical symbols, ink drop traces, folding marks, etc.). For this reason, we proceed to correct manually such error for obtaining valid wall ground truth.


\section{U-net Deep Neural Network model for wall detection}
\label{sec:unet_wall}

\begin{figure*}[!h]
\centering
\includegraphics[width=1.0\textwidth]{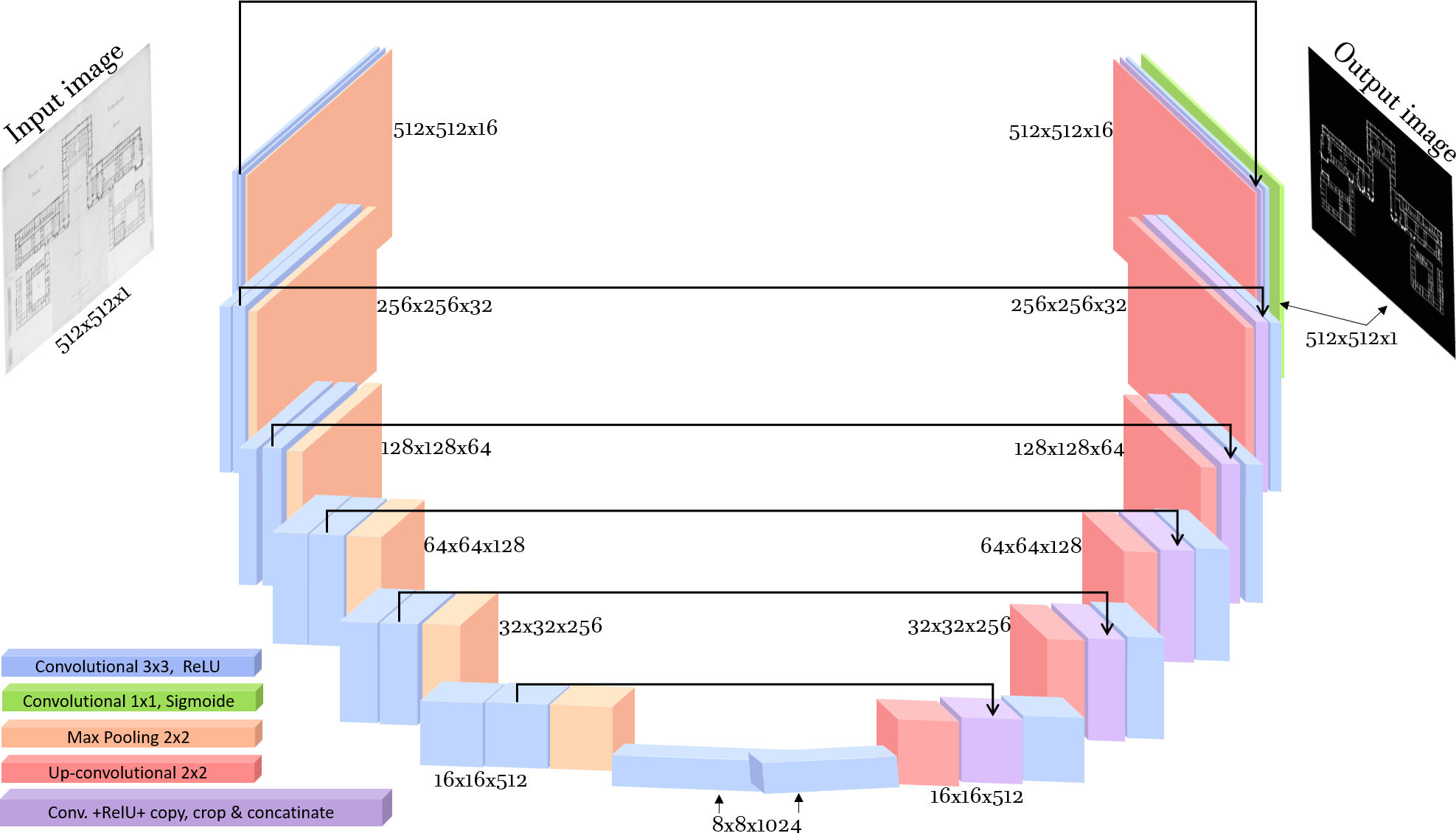}
\caption{Illustration of the used U-net architecture.}
\label{unets}
\end{figure*}

In this section, we describe another major contribution of the paper: we propose to use a convolutional neural network framework for wall detection task, we provide experimental validation of our approach in both proposed ancient architecture dataset and state-of-the-art modern datasets.
To train the proposed CNN model, we use sequential training strategy that relies on data augmentation which efficiently use the available annotated samples.
In Figure \ref{unets}, the architecture of our model is illustrated.
Our CNN model is derived from U-net neural network ~\cite{ronneberger2015u}. Our choice is motivated from the perceived advantages of U-net in case of medical image processing ~\cite{gao2017pixel} vis-a-vis speed and scarcity of training data. 


Gray scale floor plans images are scaled to $512 \times 512$ pixels of size at the input of the U-net model.
The U-net model consists of two concatenated and symmetric paths; the contracting path and the expanding path. 
The model uses the contracting path to capture the context and semantic information and the expanding path to achieve precise localization, by copying, cropping and concatenating image information observed in the contracting path toward the expanding path.
Data augmentation is often used by researchers to obtain better generalization and performance improvements specially in CNN based models..

We have used the dice loss function as introduced by Fausto, et al. \cite{milletari2016fully}.
In this study, we used the dice coefficient as a segmentation performance evaluation metric in one hand and in the other hand we used it to compute the loss function. 
The dice coefficient $D_{coef}$ is also called \textit{F-score} and calculated as shown in equation \ref{dice_c}.
\begin{equation}
     D_{coef} = \frac{2 \times |Y \cap \hat{Y}|}{|Y| + |\hat{Y}|}
\label{dice_c}
\end{equation}

Where $Y$ is the neural network prediction about the segmentation and $\hat{Y}$ is the ground truth value.
Equation \ref{dice} shows that if the prediction value is close to the ground truth value while the dice coefficient is close to $1$.
The dice loss function is defined with regards to the dice coefficient by the following equation:
\begin{equation}
     \ell_D = 1- \frac{2 \times |Y \cap \hat{Y}|+1}{|Y| + |\hat{Y}|+1}
     = 1 - D_{coef}
\label{dice}
\end{equation}

 In \cite{ronneberger2015u} authors also advocated the use of data augmentation in case of U-net based architecture
In this paper we also used a similar strategy and performed data augmentation technique to train our model.

\subsection{Training Strategy}
We use a two step learning process: First
step consists of learning the model on the modern CVC-FP dataset samples \cite{de2015cvc}. As the Versailles palace dataset samples have quite different background than that of CVC-FP, we replaced the white background in the CVC original samples by Versailles colored plans background. This step provides a good pre-training for our final model.
In fact we will see (in experiments section) this initial model obtains quite impressive result

The choice of the learning rate, batch size and number of epochs hyper parameters in addition to the most convenient optimisation algorithm and loss function is empirically decided during this step of the model training and used for the rest of the training.

Next, We obtain our final model by learning on Versailles dataset samples, however instead of initialize the parameters randomly, we use the parameters from the model trained on CVC-FP in first step.
\subsection{Training and Implementation Details}
Best performance of the initial model on CVC-FP dataset is obtained with learning rate of $10e^{-4}$, batch size of $16$ with $100$ training epochs with active early stopping after $50$ epochs. 
The model learned with $RMSprop$ optimizer with the dice loss function $\ell_D$ outperforms the models learned with SGD, Adam or Ftrl optimizers and binary cross-entropy, weight cross-entropy or focal loss functions.
In the first step of the sequential training strategy we followed, an initial U-net model is trained on the modified CVC-FP dataset with Versailles-FP background. we step up the training for 100 epochs with early stopping after 50 epochs. 
The training data are augmented five times its size divided by the batch size where the batch size is equal to 16.
We observed that the most effective data augmentation settings are zoom=0.3, right/left shift=0.2 with activated horizontal and vertical flipping.

\section{Experiments}
\label{sec:dl_exper}

In this section, we present the experiments carried out to validate our proposals regarding automatic mask generation and U-net based wall detection .
In the following we first describe in Section \ref{sec:eval_protocol} different experimental protocols that are followed for different datasets and then provide experimental validation for both mask generation in Section \ref{sec:mask_gen} and wall detection in Section \ref{sec:wall_detection}. 


\subsection{Dataset and evaluation protocol}
\label{sec:eval_protocol}
\begin{figure*}[!h]
\centering
\includegraphics[width=0.8\textwidth]{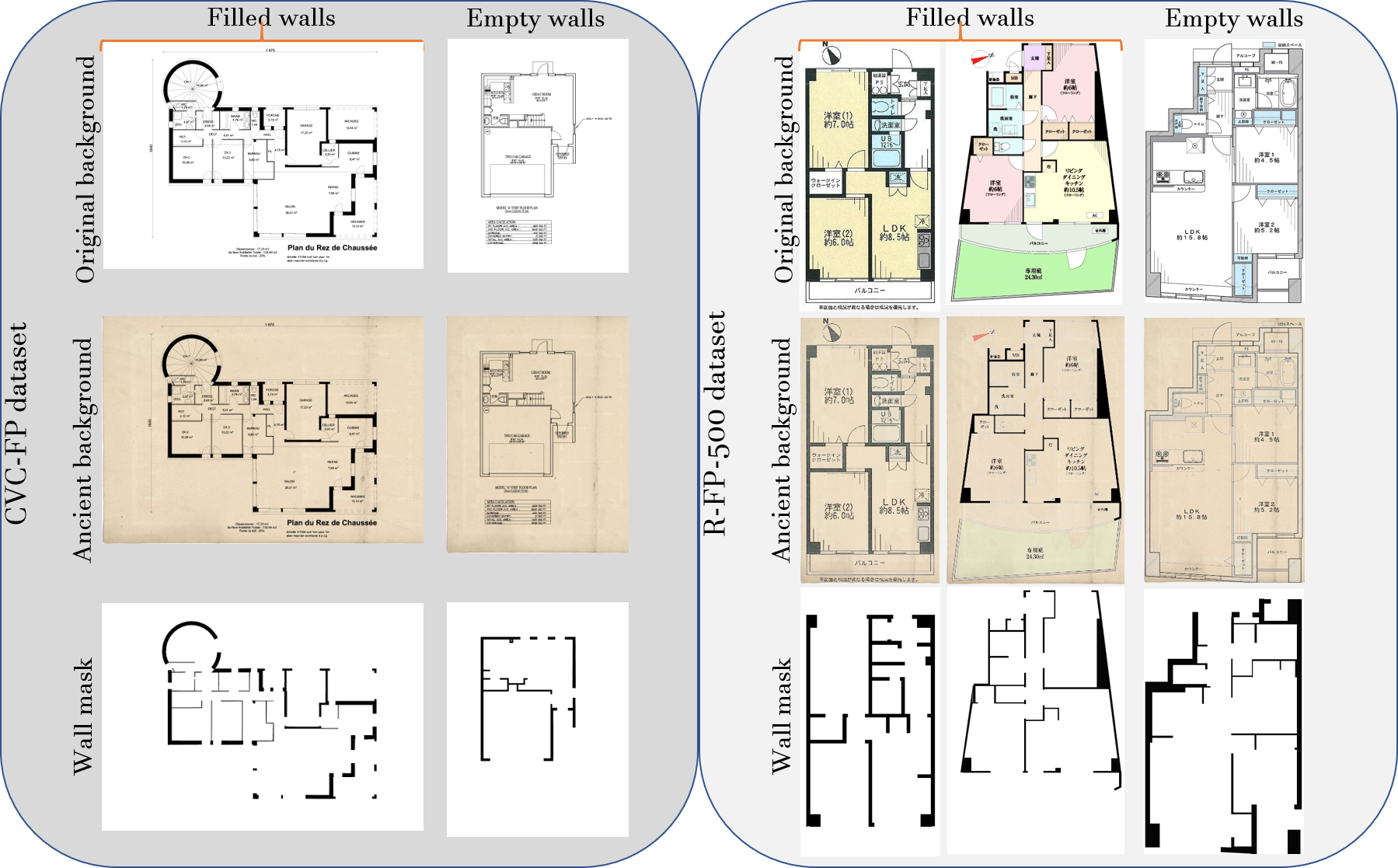}
\caption{Samples of CVC-FP and R-FP-500 data set; first image row show the original floor plan, second row show the floor plans with Versailles background and last row show the wall masks.}
\label{db}
\end{figure*}

\subsubsection{Dataset:}
Only a few modern floor plan datasets are available for research due to the requirement of challenging collaborative work. 
In our experiments, we considered the CVC-FP \cite{de2015cvc}  and R-FP \cite{DodgeMVA2017} modern and real floor plan datasets for evaluating the performance of the proposed wall mask generation approach and the U-net wall segmentation model.
CVC-FP dataset is a collection of 122 real floor plan scanned documents compiled and groundtruthed using SGT toolkit\footnote{\url{http://dag.cvc.uab.es/resources/floorplans/}}.
The dataset consists in four sets of plans that have different drawing styles, image quality and resolution incorporating graphics, text descriptions and tables. 
R-FP dataset \cite{DodgeMVA2017} is collected from a public real estate website.
It consists of 500 floor plan images that varies in size and contains different sort of information, decorative elements, shading schemes and colored backgrounds.

To the best of our knowledge, there is no available annotated dataset for ancient floor plans.
We introduce in this study, first ancient floor plan dataset, named Versailles-FP dataset which is based on the french VERSPERA research project. 
The dataset is collected from scanned floor plans that belongs to Versailles palace constructions dated of $17^{th}$ and $18^{th}$ century. 
We used the proposed wall mask generation approach (Section \ref{sec:mask_gen}) to annotate 500 images of Versailles palace floor plans.
Figure \ref{dataset} shows samples of this dataset.

\subsubsection{Evaluation Protocol - Mask Generation:}
We evaluated the performance of the proposed wall mask generation approach on modified CVC-FP and R-FP datasets with Versailles-FP background. Mean dice coefficient and mean IoU metrics are used as evaluation criteria.
In order to critically analyze we further split every dataset into two
evaluation sets based on wall types 1) filled wall evaluation set that contains floor plan images with
filled walls. 2) empty wall evaluation set where the walls in the floor plan images
are empty.

\subsubsection{Evaluation Protocol - Wall Detection:}
We conducted experiments and reported results on the following 5 datasets ; 1) original CVC-FP dataset, 2) CVC-FP with Versailles-FP background dataset, 3) original R-FP dataset, 4) R-FP with
Versailles-FP  background  dataset  and  5)  Versailles-FP  dataset. 
For these five datasets, we employed a 5-fold cross validation scheme, where each datasets is split into 3 parts: taking $3/5$ samples for training, $1/5$ samples for validating and rest $1/5$ samples for testing.
The final result is calculated by taking mean over all test folds for each datasets.
As evaluation metrics, we use the Dice coefficient defined in equation \ref{dice} and mean Intersection-over-Union (IoU) as in  \cite{DodgeMVA2017} and \cite{long2015fully}.

\subsection{Wall mask generation evaluation}
\label{sec:mask_gen}
 
To validate our semi-automatic ground truth generation procedure, we introduce this pseudo-task to analyze the efficacy and correctness of our approach, which can provide an idea about the efficacy of our approach that can be utilized further to expedite generation of such datasets. 
\begin{table}[!h]
\caption{Evaluation results of filled and empty Wall's masks generation}
\label{tab:my-table}
\resizebox{\textwidth}{!}{%
\begin{tabular}{l|c|c|c|c|}
\cline{2-5}
 &
  \multicolumn{2}{c|}{\begin{tabular}[c]{@{}c@{}}CVC dataset\\ (122 FP-image)\end{tabular}} &
  \multicolumn{2}{c|}{\begin{tabular}[c]{@{}c@{}}R-FP dataset\\ (500 FP-image)\end{tabular}} \\ \cline{2-5} 
 &
  \begin{tabular}[c]{@{}c@{}}Filled Walles\\ (70\% of the dataset)\end{tabular} &
  \begin{tabular}[c]{@{}c@{}}Empty Walls\\ (30\% of the dataset)\end{tabular} &
  \begin{tabular}[c]{@{}c@{}}Filled Walls\\ (73\% of the dataset)\end{tabular} &
  \begin{tabular}[c]{@{}c@{}}Empty Walls\\ (27\% of the dataset)\end{tabular} \\ \hline
\multicolumn{1}{|l|}{Dice score \%} &
  90.75\% &
  34.45\% &
  85.09\% &
  26.20\% \\ \hline
\multicolumn{1}{|l|}{IoU score \%} &
  83.05\% &
  21.78\% &
  76.36\% &
  19\% \\ \hline
\end{tabular}%
}
\label{mas}
\end{table}

Table \ref{mas} introduce the evaluation results on the filled and empty walls evaluation sets of CVC-FP and R-FP dataset.
We can observe from figure \ref{db} that wall masks are always filled masks even for hollow (empty) walls in the the floor plan images, we consider this as an annotation error.
However, the proposed approach produce wall mask identical to the walls in the input image. 
Due to this, we obtained very low dice and IoU scores when evaluating on hollow(empty) walls evaluation sets.

On the other hand, the proposed mask generation approach performs well on $70\%$ of CVC-FP and $73\%$ of R-FP-500 plan images that belongs to the filled walls evaluation sets.  
From table \ref{mas}, we observe that approach achieves $90.75\%$ and $83.05\%$ of dice coefficient and IoU scores on the filled walls evaluation set of CVC-FP dataset. 
At same time, we observe a performance degradation on the filled walls of the R-FP-500 dataset measured by $5.66\%$ and $6.69\%$ of dice and IoU scores respectively.
This degradation effect can be explained by the multi-color backgrounds of the R-FP-500 dataset images.
By analysing the masks that are generated by the approach, we observed that the approach successes in eliminating most of texts, description tables and decoration elements but failed in case of orientation sign graphics and trees.

As this step can be used for generation of groundtruth, it is worth mentioning that our approach takes only around 3.5 minutes (not considering the time taken for manual correction ) for one image of size $17000 \times 8000$ in comparison to 30-35 minutes on average by human annotator. Thus achieving 10X speed-up in annotation process.

\subsection{Analysis of Wall Detection Experiment}
\label{sec:wall_detection}
In wall detection we have obtaned the model in two steps, first we pretrain the proposed U-net model by training only on modified CVC-FP dataset and this alone produce impressive results for CVC-FP dataset.
Our model achieve 99.57\% accuracy and can separate walls with considerable difficulties.

Next, we further train our detection model on 5 different datasets, where parameters are initialized using the pretrained model in previous step. The result of this model is provided in Table \ref{tab:results}.

\begin{table}[!h]
\centering

\caption{Wall segmentation evaluation and comparison with the state-of-the-art }
\label{tab:results}
\resizebox{\textwidth}{!}{%
\tiny
\begin{tabular}{|l|l|l|l|l|}
\hline
Dataset         & Model  & Mean accuracy \% & Mean IoU \% & Mean Dice \% \\ \hline
Original CVC-FP & FCN-2s \cite{DodgeMVA2017} & 97.3             & 94.4        & NA           \\ \hline
Original CVC-FP & U-net  & 99.73            & 95.01       & 95.65        \\ \hline
Modified CVC-FP & U-net  & 99.71            & 93.45       & 95.01        \\ \hline
Original R-FP   & FCN-2s \cite{DodgeMVA2017} & 94.0             & 89.7        & NA           \\ \hline
Original R-FP   & U-net  & 98.98            & 90.94       & 91.29        \\ \hline
Modified R-FP   & U-net  & 98.92           & 90.21       & 91.00        \\ \hline
Versailles-FP   & U-net  & 97.15            & 88.14       & 93.32        \\ \hline
\end{tabular}%
}
\end{table}

It can be observed from table\ref{tab:results} that our U-net model outperform state-of-the-art FCN-2s model \cite{DodgeMVA2017} on the modern floor plan datasets of CVC-FP and R-FP-500 with original (white) background.
Thus providing a better state-of-the-art result in wall detection task.
It should also be noted that the proposed convolutional model is more efficient, than a fully connected model as it uses fewer weights, consequently  the proposed model should be faster in respect to compuational speed.
\section{Conclusion}
\label{sec:conclusion}
In this paper we introduced the Versailles-FP dataset that consists on 500 ancient floor plan images and wall masks of The Versailles Palace dated of $17^{th}$ and $18^{th}$ century. 
At first, the wall masks are generated automatically using multi-orientation steerable filters based approach. Then, the generated wall masks are validated and corrected manually. We showed that the mask generator performs well enough on the filled walls of CVC-FP and R-FP datasets. A U-net architecture is then used to achieve the wall detection task on CVC-FP, R-FP and Versailles-FP datasets. The cross-validation results confirmed that the convoulutional architecture of the U-net model performs better than the fully connected network architecture.

\subsection*{Acknowledgements}
We thank our colleagues of the VERSPERA research project in the Research Center of Château de Versailles, French national Archives and French national Library, and the Fondation des sciences du patrimoine which supports VERSPERA.

\bibliographystyle{splncs04}

\bibliography{samplepaper}

\end{document}